\documentclass[11pt]{article}

\usepackage{amsmath}
\usepackage{amssymb}
\usepackage{booktabs}
\usepackage{graphicx}
\usepackage[colorlinks=true,linkcolor=blue,citecolor=blue,urlcolor=blue]{hyperref}
\usepackage[round]{natbib}
\usepackage[margin=1in]{geometry}
\usepackage{xcolor}
\usepackage{multirow}
\usepackage{placeins}

\title{Beyond Expression Similarity: Contrastive Learning Recovers Functional Gene Associations from Protein Interaction Structure}
\author{Jason Dury\\Eridos AI\\jason@eridos.ai}
\date{}

\begin{document}

\maketitle

\begin{abstract}
\noindent The Predictive Associative Memory (PAM) framework posits that useful relationships often connect items that co-occur in shared contexts rather than items that appear similar in embedding space.\footnote{Code and data: \url{https://github.com/EridosAI/GeneticCAL}} A contrastive MLP trained on co-occurrence annotations---the procedure we term Contrastive Association Learning (CAL)---has been shown to improve multi-hop passage retrieval and discover narrative function at corpus scale in text. We test whether this principle transfers to molecular biology, where protein--protein interactions provide independently derived functional associations distinct from gene expression similarity.

Four experiments across two biological domains map the operating envelope. On gene perturbation data (Replogle K562 CRISPRi, 2,285 genes), CAL trained on STRING protein interactions achieves cross-boundary AUC of 0.908 where expression similarity scores 0.518. A second gene dataset (DepMap, 17,725 genes) confirms the result after a negative sampling correction, reaching cross-boundary AUC of 0.947. Two drug sensitivity experiments produce informative negatives that sharpen the boundary conditions. Three cross-domain findings emerge: (1)~inductive transfer succeeds in biology---a node-disjoint split with completely unseen genes yields AUC 0.826 ($\Delta$ +0.127)---where it fails in text ($\pm 0.10$), suggesting physically grounded associations are more transferable than contingent co-occurrences; (2)~CAL association scores anti-correlate with interaction degree (Spearman $r = -0.590$), with cross-boundary gains concentrating on understudied genes with focused interaction profiles; (3)~tighter association quality outperforms larger but noisier training sets, reversing the pattern observed in text. Results on the primary dataset are stable across training seeds (SD $< 0.001$) and cross-boundary threshold choices.
\end{abstract}

\section{Introduction}
\label{sec:introduction}

A kinase and the transcription factor it phosphorylates play fundamentally different roles in the same signalling pathway. Knocking out one produces a different downstream expression signature than knocking out the other. In embedding space, their gene perturbation profiles may be distant---they do not look alike. Yet they are functionally associated: they participate in the same biological process, and understanding one requires knowing the other.

This pattern was first characterised in text retrieval, where passages required for the same reasoning chain are often distant in embedding space. The Predictive Associative Memory (PAM) framework \citep{dury2026pam} formalised the distinction, and a contrastive MLP trained on co-occurrence annotations---Contrastive Association Learning (CAL)---improved multi-hop passage retrieval by +8.6 Recall@5 \citep{dury2026aar} and discovered clusters organised by narrative function rather than topic at corpus scale \citep{dury2026concepts}..

These results establish that association and similarity produce different retrieval behaviour in text. But is this a property of language specifically, or a general phenomenon? Text has particular characteristics that might explain the results: natural language is high-dimensional and richly structured, and the co-occurrence annotations in multi-hop QA datasets reflect human judgements about reasoning chains. If association $\neq$ similarity holds only in this regime, the theoretical claim is narrower than proposed.

Molecular biology provides a test case with several desirable properties. First, functional associations exist independently of the embeddings: the STRING database \citep{szklarczyk2023string} catalogues protein--protein interactions supported by experimental evidence (co-immunoprecipitation, affinity purification, yeast two-hybrid), text-mining, co-expression, and other channels. Second, gene perturbation profiles from CRISPRi screens \citep{replogle2022mapping} provide embeddings that measure the downstream effect of gene knockdown---a fundamentally different information source than the interaction annotations. Third, a cross-boundary regime exists: many interacting protein pairs have low expression similarity because functionally related genes can play different roles in the same pathway.

We apply CAL to four biological datasets spanning two domains (gene perturbation, drug sensitivity). The same 4-layer MLP architecture, the same symmetric InfoNCE loss, and the same evaluation framework transfer without modification beyond input dimensionality.

Two experiments produce strong positive results. Two produce informative negatives. Together, they map the operating envelope of contrastive association learning and reveal three domain differences that the text experiments alone could not have identified: an inductive transfer asymmetry that distinguishes physically grounded from contingent associations, a degree-dependence pattern that concentrates CAL's utility on understudied genes, and a quality-over-quantity dynamic that reverses the text domain's preference for more training pairs.

\section{Background}
\label{sec:background}

\subsection{Contrastive Association Learning}

CAL trains a function $f: \mathbb{R}^d \to \mathbb{R}^d$ to map embeddings into an association space where co-occurring items are close and non-co-occurring items are distant. The architecture is a 4-layer MLP with LayerNorm, GELU activations, and a learned residual connection:
\begin{equation}
f(\mathbf{x}) = \text{normalize}\bigl(\alpha \cdot \mathbf{x} + (1 - \alpha) \cdot g(\mathbf{x})\bigr)
\label{eq:residual}
\end{equation}
where $g$ is the MLP transformation (three hidden layers of dimension 1024), $\alpha = \sigma(\alpha_{\text{logit}})$ is a learned scalar controlling the residual blend, and the output is L2-normalised. The model has approximately 2.2M parameters.

Training uses symmetric InfoNCE loss: for a batch of $B = 512$ pairs, compute the cosine similarity matrix between transformed embeddings and apply cross-entropy loss in both directions, averaged:
\begin{equation}
\mathcal{L} = \frac{1}{2}\bigl[\text{CE}(\mathbf{S}, \mathbf{y}) + \text{CE}(\mathbf{S}^\top, \mathbf{y})\bigr]
\label{eq:loss}
\end{equation}
where $S_{ij} = f(\mathbf{e}_{a_i})^\top \cdot \mathbf{e}_{b_j} / \tau$, $\tau = 0.05$ is the temperature, and $\mathbf{y} = (0, 1, \ldots, B-1)$ are diagonal targets. Note the asymmetry: the first element of each pair is transformed through $f$ while the second is used as a raw embedding. Optimisation uses AdamW (lr $= 3 \times 10^{-4}$, weight decay $= 1 \times 10^{-4}$) with cosine annealing over 100 epochs.

This contrastive training procedure was introduced in \citet{dury2026pam} for learning temporal associations in synthetic environments, applied to multi-hop passage retrieval in \citet{dury2026aar}, and used for corpus-scale concept discovery in \citet{dury2026concepts}. In text retrieval, CAL improved Recall@5 by +8.6 points on HotpotQA in the transductive setting while an inductive variant showed no significant improvement ($\pm 0.10$), and training on semantically similar but non-associated pairs degraded performance below the cosine baseline \citep{dury2026aar}.

The present work applies CAL to biological data with zero architectural modification beyond input dimensionality (50 dimensions from PCA-reduced gene expression versus 1024 from BGE-large-en-v1.5 text embeddings).

\subsection{Biological Setup}
\label{sec:bio_setup}

The biological analogue maps directly onto the text framework:

\begin{itemize}
    \item \textbf{Embeddings:} Gene perturbation profiles from CRISPRi screens, where each gene's embedding is the downstream expression change caused by its knockdown. These are measured signals, not learned representations. Profiles are PCA-reduced to 50 dimensions and L2-normalised.
    \item \textbf{Associations:} Protein--protein interactions from the STRING database. STRING provides a combined confidence score aggregating multiple evidence channels including experimental assays, text-mining, co-expression, and curated databases. We filter on this combined score at three thresholds ($\geq$400, $\geq$700, $\geq$900) and validate separately against the experimental evidence channel alone (Section~\ref{sec:quality}).
    \item \textbf{Cross-boundary regime:} Interacting proteins can have dissimilar perturbation profiles because they play different functional roles in the same pathway. We define cross-boundary pairs as those with $|\text{cosine similarity}| < 0.2$, the regime where cosine-based discrimination drops to chance.
\end{itemize}

\subsection{Evaluation}
\label{sec:evaluation}

At evaluation time, we score each gene pair using half-transformed association scoring:
\begin{equation}
\text{score}(A, B) = \frac{1}{2}\bigl(f(\mathbf{e}_A) \cdot \mathbf{e}_B + f(\mathbf{e}_B) \cdot \mathbf{e}_A\bigr)
\label{eq:scoring}
\end{equation}
where $\mathbf{e}_A$ and $\mathbf{e}_B$ are the L2-normalised PCA-50 embeddings. This mirrors the bi-directional scoring used in text retrieval \citep{dury2026aar}.

The primary metric is AUC for the binary discrimination task: given a pair of genes, are they STRING-associated or not? Negative pairs are sampled uniformly at random from all non-associated gene pairs ($5\times$ the number of positives, capped at 50,000; seed 42). We additionally report cross-boundary AUC restricted to pairs with $|\text{cosine}| < 0.2$, isolating the regime where embedding similarity is uninformative.

On the primary Replogle dataset, all results report association-only scoring (no cosine blending). The learned association function monotonically outperforms any blend with cosine similarity across all thresholds (full sweep in Appendix~\ref{app:lambda}), so blending with cosine only dilutes the signal. This contrasts with text retrieval, where the optimal blend ($\lambda = 0.6$) balances cosine and association \citep{dury2026aar}.

Results are reported as mean $\pm$ SD across training seeds (42, 123, 456) where indicated. All other results use seed 42.

\subsection{Relationship to Existing Work}
\label{sec:related}

CAL is not proposed as a protein--protein interaction prediction method. Established approaches to PPI prediction---including graph neural networks on interaction networks, node2vec embeddings, and supervised link prediction---are designed and optimised for that task. The biological experiments here validate the cross-domain generality of the association $\neq$ similarity principle, with STRING interactions providing reference associations for evaluation rather than representing the target application. We compare against cosine similarity as the baseline because the claim under test is whether learned associations capture information that embedding similarity misses, not whether CAL outperforms dedicated PPI prediction pipelines.

\section{Experiment 1: Gene Perturbation Profiles $\times$ Protein Interactions}
\label{sec:replogle}

\subsection{Dataset}

We use the \citet{replogle2022mapping} K562 essential CRISPRi Perturb-seq dataset: 2,285 gene perturbations measured across 8,563 gene expression readouts. Embeddings are PCA-50 projections of the perturbation profiles, capturing approximately 50\% of variance, L2-normalised before training. Mean pairwise cosine similarity is 0.088, indicating a broad, relatively unstructured embedding space.

Association pairs are drawn from STRING v12.0 \citep{szklarczyk2023string} protein--protein interactions, filtered on combined confidence score. Gene mapping coverage is 98.1\% (2,019 of 2,058 perturbation genes matched). We evaluate at three confidence thresholds:

\begin{table}[h]
\centering
\caption{STRING pair statistics at each confidence threshold.}
\label{tab:dataset}
\begin{tabular}{lrrcc}
\toprule
Threshold & Pairs & CB ($|$cos$| < 0.2$) & Mean pos.\ cosine & Frac.\ cos $> 0.5$ \\
\midrule
Low ($\geq$400) & 93{,}562 & 34.9\% & 0.187 & 26.3\% \\
Medium ($\geq$700) & 41{,}434 & 25.6\% & 0.306 & 39.9\% \\
High ($\geq$900) & 23{,}268 & 20.2\% & 0.389 & 50.2\% \\
\bottomrule
\end{tabular}
\end{table}

Even at the highest confidence, 49.8\% of associated pairs have cosine similarity $\leq 0.5$ and 20.2\% fall in the cross-boundary regime, a substantial fraction of real functional associations that cosine similarity cannot identify.

\subsection{Main Results}

\begin{table}[h]
\centering
\caption{Confidence threshold sweep (Replogle K562 + STRING). See also Figure~\ref{fig:overview}.}
\label{tab:main}
\begin{tabular}{lrrr}
\toprule
Metric & Low ($\geq$400) & Medium ($\geq$700) & High ($\geq$900) \\
\midrule
Training pairs & 93{,}562 & 41{,}434 & 23{,}268 \\
Cosine baseline AUC & 0.570 & 0.644 & 0.692 \\
CAL AUC (assoc-only) & 0.800 & 0.877 & \textbf{0.910} \\
$\Delta$AUC & +0.230 & +0.233 & +0.218 \\
\midrule
CB cosine AUC & 0.529 & 0.534 & 0.518 \\
CB CAL AUC & 0.783 & 0.861 & \textbf{0.908} \\
$\Delta$CB AUC & +0.254 & +0.327 & +0.390 \\
\bottomrule
\end{tabular}
\end{table}

At high confidence ($\geq$900), CAL achieves cross-boundary AUC of 0.908, which is strong discrimination between interacting and non-interacting gene pairs in the regime where cosine similarity is at chance (0.518). Results are stable across training seeds: overall AUC $0.911 \pm 0.0005$, cross-boundary AUC $0.908 \pm 0.0006$ (seeds 42, 123, 456).

Cross-boundary AUC is robust to threshold choice: at $|\text{cos}| < 0.30$, $< 0.20$, $< 0.15$, $< 0.10$, and $< 0.05$, CAL AUC remains approximately 0.91 throughout (full table in Appendix~\ref{app:cb}).

\begin{figure}[t]
\centering
\includegraphics[width=\textwidth]{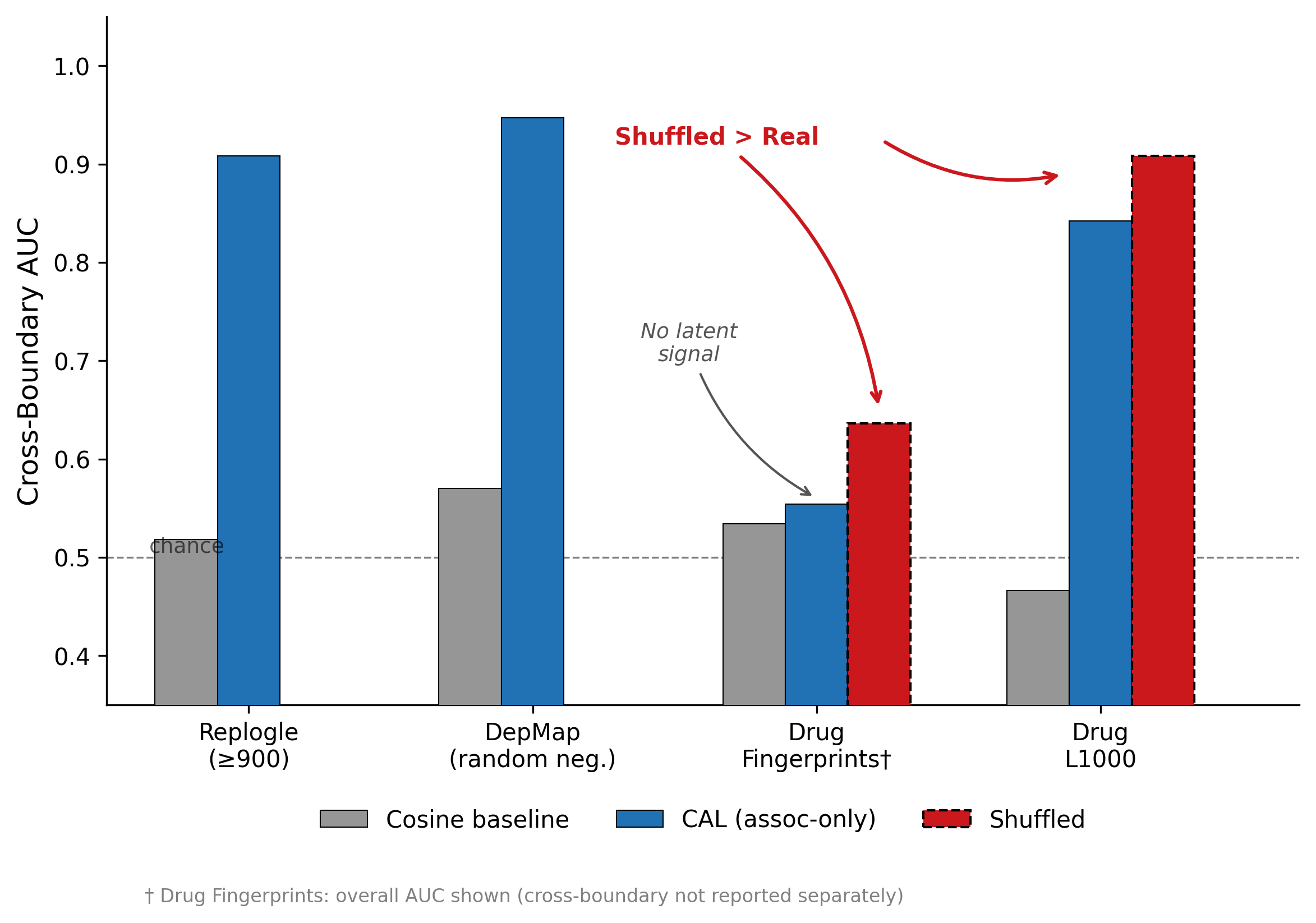}
\caption{Cross-boundary AUC across all four experiments. CAL provides large gains on gene perturbation data where latent associative signal exists (Replogle, DepMap). Drug experiments reveal failure modes: no latent signal (fingerprints) and degree confounding (L1000, where shuffled $>$ real).}
\label{fig:overview}
\end{figure}

\subsection{Quality Over Quantity}
\label{sec:quality}

The monotonic improvement with tighter thresholds---despite fewer training pairs---reverses the pattern observed in text, where more co-occurrence pairs improved performance \citep{dury2026aar}. STRING's combined score aggregates multiple evidence channels. Low-confidence edges are more likely to be driven by text-mining and co-expression, which are themselves similarity signals. Higher-confidence edges have stronger aggregate evidence, including more experimental support.

To validate this interpretation, we trained a separate model filtering exclusively on the experimental evidence channel (ignoring text-mining, co-expression, and other channels). Experimental-only filtering at $\geq$700 produces 0.907 overall AUC---approaching the combined-score result at $\geq$900 (0.910) despite using a different and smaller set of pairs. Both use 23,000+ pairs, but drawn from different filtering criteria. This confirms that the quality effect is driven by the strength of physical interaction evidence, not simply by the combined score being higher.

\begin{table}[h]
\centering
\caption{Combined-score vs experimental-channel filtering. See also Figure~\ref{fig:threshold}.}
\label{tab:expstring}
\begin{tabular}{lrrr}
\toprule
Filter & Pairs & Overall AUC & CB AUC \\
\midrule
Combined $\geq$700 & 41{,}434 & 0.877 & 0.861 \\
Combined $\geq$900 & 23{,}268 & \textbf{0.910} & \textbf{0.908} \\
Experimental $\geq$700 & 23{,}438 & 0.907 & 0.902 \\
\bottomrule
\end{tabular}
\end{table}

\begin{figure}[t]
\centering
\includegraphics[width=\textwidth]{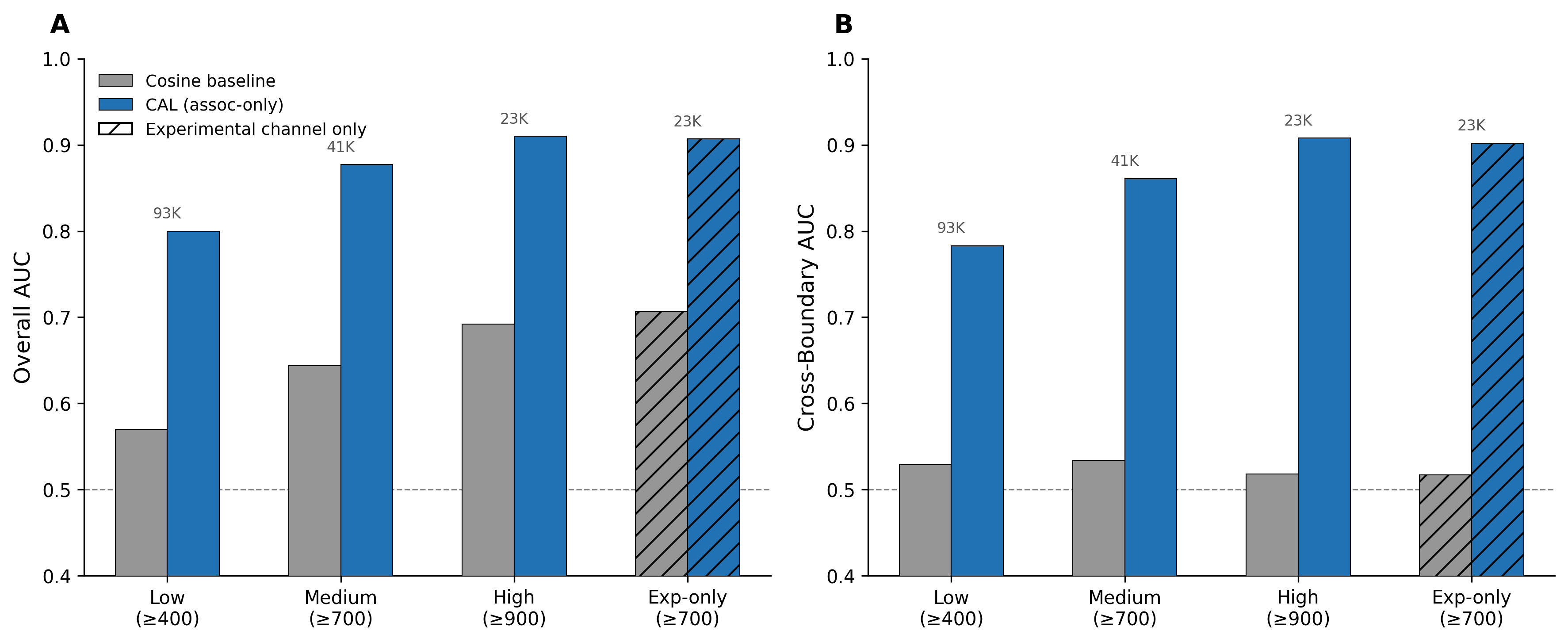}
\caption{Confidence threshold sweep showing quality-over-quantity. (A)~Overall AUC and (B)~cross-boundary AUC both improve with tighter thresholds despite fewer training pairs. Experimental-only filtering (hatched) matches the highest combined-score threshold.}
\label{fig:threshold}
\end{figure}

\subsection{Ablations}
\label{sec:ablations}

\begin{table}[h]
\centering
\caption{Ablation study (Replogle, medium confidence $\geq$700).}
\label{tab:ablations}
\begin{tabular}{lrl}
\toprule
Condition & $\Delta$AUC vs cosine & Interpretation \\
\midrule
Full CAL & +0.233 & Full model \\
Shuffled pairs & $-$0.096 & Random pairings actively hurt \\
Similar positives & +0.018 & Association $\neq$ similarity confirmed \\
Random negatives & +0.193 & Positives carry signal \\
Inductive (edge-split 70/30) & +0.183 & Compositional generalisation \\
Inductive (node-split 70/30) & +0.127 & Cold-start generalisation (Section~\ref{sec:inductive}) \\
\bottomrule
\end{tabular}
\end{table}

All four ablations behave as predicted from the text experiments:

\textbf{Shuffled pairs ($-$0.096):} Randomly permuting the association pairs produces a model that performs worse than cosine similarity alone. At association-only scoring, random pairings actively hurt---the model learned genuine STRING interaction structure, not artifacts of the training procedure.

\textbf{Similar positives (+0.018):} Training on the highest-cosine gene pairs adds almost nothing beyond the cosine baseline. Learning ``which genes have similar expression'' does not help. This is the biological instantiation of the text finding that training on similar-but-not-associated passages degrades retrieval \citep{dury2026aar}.

\textbf{Random negatives (+0.193):} Replacing in-batch negatives with randomly sampled negatives yields 0.837, which is 0.040 below the full model. As in text, the positive pairs carry the primary training signal. In-batch negatives work here because Replogle positive pairs are spread across the cosine range (only 39.9\% have cosine $> 0.5$ at medium confidence), unlike DepMap where positives cluster (Section~\ref{sec:depmap}).

\textbf{Inductive:} Discussed in Section~\ref{sec:inductive}.

\section{Experiment 2: Gene Expression $\times$ Co-Essentiality}
\label{sec:depmap}

\subsection{Dataset}

The DepMap project \citep{dempster2019depmap} provides CRISPR-Cas9 knockout screening across 1,112 cancer cell lines. We use 17,725 genes with RNA-seq expression profiles (PCA-100, 94.7\% variance) as embeddings and co-essentiality scores derived from the knockout screens as the association signal.

\subsection{The Negative Sampling Problem}

Co-essential genes are overwhelmingly co-expressed: 98\% of the top co-essential pairs have expression cosine $> 0.5$, and the cosine baseline AUC is already 0.896. With in-batch negatives, the 511 negative examples per positive are themselves co-essential genes from the same expression cluster. The model learns to push apart genes that should be together.

\begin{table}[h]
\centering
\caption{DepMap training results: in-batch vs random negatives.}
\label{tab:depmap}
\begin{tabular}{lrrrr}
\toprule
Setting & N pairs & CAL AUC & Cosine AUC & $\Delta$AUC \\
\midrule
In-batch negatives & 200K & 0.379 & 0.896 & $-$0.517 \\
Random negatives & 200K & \textbf{0.956} & 0.896 & \textbf{+0.060} \\
\bottomrule
\end{tabular}
\end{table}

With in-batch negatives, CAL anti-discriminates at all training set sizes tested (25K--200K pairs; Appendix~\ref{app:depmap}). Switching to random negatives---drawing negatives uniformly from all gene pairs rather than from the batch---rescues performance completely: overall AUC 0.956, cross-boundary AUC 0.947 (cosine: 0.570), cross-boundary at $|\text{cos}| < 0.1$: AUC 0.946 (cosine: 0.539). DepMap results use both-transformed scoring ($f(\mathbf{e}_A) \cdot f(\mathbf{e}_B)$) from the original experiment, while Replogle results use half-transformed scoring (Section~\ref{sec:evaluation}). Both methods capture the same qualitative pattern; the choice affects absolute values but not the direction or magnitude of the CAL advantage.

The failure diagnosis is straightforward: in-batch negatives fail when positive pairs cluster in embedding space. The Replogle dataset does not have this problem (only 39.9\% of positive pairs have cosine $> 0.5$ at medium confidence), which is why in-batch negatives work there. Before applying CAL to new data, check the cosine distribution of positive pairs. Clustered positives require random negatives.

External validation against STRING interactions (independent of the co-essentiality training signal) shows that the DepMap model's overall STRING AUC (0.616--0.695) falls below cosine (0.651--0.712), consistent with the both-transformed scoring function not being optimised for this task. However, cross-boundary AUC---the regime where cosine is uninformative---shows modest but consistent improvement: CAL 0.537--0.552 vs cosine 0.516--0.522 across STRING confidence thresholds.

\section{Experiments 3 and 4: Drug Sensitivity}
\label{sec:drugs}

Two drug experiments probe the boundaries of CAL by applying it to drug co-lethality in cancer cell lines, using the PRISM repurposing screen \citep{corsello2020prism}.

\subsection{Experiment 3: Morgan Fingerprints $\times$ Co-Lethality}

Drug structure embeddings (2048-bit Morgan fingerprints from SMILES strings, 4,684 drugs) have mean pairwise cosine of 0.180, with 64\% of co-lethal pairs in the cross-boundary regime. The cosine baseline AUC is 0.534---barely above chance. These statistics initially appeared well-suited for CAL.

Training fails. At 200K pairs, CAL AUC reaches only 0.554 with training accuracy stuck at 0.2\%. The shuffled ablation is diagnostic: shuffled AUC (0.636) exceeds the reference (0.554), indicating the model learned degree structure rather than co-lethality. External validation confirms: cosine similarity alone better recovers mechanism of action (0.716 vs 0.553) and drug target (0.772 vs 0.582). Full results in Appendix~\ref{app:drugs}.

In this setup, Morgan fingerprints contain no usable latent signal for co-lethality. Co-lethality depends on 3D binding geometry, target selectivity, and pharmacokinetics---information not accessible from radius-2 atomic substructure counts. The embedding and association appear genuinely orthogonal.

\subsection{Experiment 4: Transcriptional Signatures $\times$ Co-Lethality}

L1000 transcriptional signatures (978 landmark genes, 2,094 drugs matched to PRISM) measure what a drug does to cells, not what it looks like. Cross-boundary fraction is 74--80\% and cosine baseline is at chance (0.466).

With random negatives, CAL reaches AUC 0.850 and cross-boundary AUC 0.842---apparently strong. The shuffled ablation kills it: shuffled AUC 0.909 exceeds the reference. With 200K pairs drawn from only 2,094 drugs, the average drug appears in $\sim$191 pairs. The model learns drug frequency, not co-lethality.

Degree-controlled analysis reveals a residual signal: among the lowest-degree drugs (Q0, 420 pairs), the reference model outperforms shuffled by +0.091 (0.904 vs 0.813), suggesting some genuine co-lethality structure exists in the low-degree regime. But the overall metric is dominated by degree confounding.

Here, low entity diversity relative to pair count allows degree signal to dominate. An entity-to-pair ratio of approximately 1:95 means degree signal overwhelms genuine association signal. This confound is detectable by the shuffled ablation.

\FloatBarrier
\section{Cross-Domain Analysis}
\label{sec:cross_domain}

The four experiments, together with the text results from \citet{dury2026aar} and \citet{dury2026concepts}, produce three findings that no single experiment could have identified.

\subsection{Inductive Transfer: Contingent vs Physically Grounded Associations}
\label{sec:inductive}

\begin{table}[h]
\centering
\caption{Inductive transfer across domains.}
\label{tab:inductive}
\begin{tabular}{llrl}
\toprule
Domain & Split type & $\Delta$ vs baseline & Notes \\
\midrule
Text (HotpotQA) & Edge-split & +0.10 R@5 & Not significant (CI includes zero) \\
Text (MuSiQue) & Edge-split & $-$7.63 R@5 & Harmful \\
Biology (Replogle, med.) & Edge-split & +0.183 AUC & Genes overlap between train/test \\
Biology (Replogle, high) & \textbf{Node-split} & \textbf{+0.127 AUC} & \textbf{Unseen genes, cold-start} \\
\bottomrule
\end{tabular}
\end{table}

The node-disjoint split is the stronger test. Thirty percent of genes are held out entirely---no pair involving these genes appears in training. The model must generalise to genes it has never seen. Test AUC reaches 0.826 ($\Delta$ +0.127 over cosine), providing evidence of cold-start generalisation. The edge-split at medium confidence retains more improvement (+0.183), suggesting an additional compositional benefit when individual gene representations are familiar. Comparing at medium confidence, where both splits are available, the node-split retains 69\% of the edge-split improvement (+0.126 vs +0.183), suggesting that much of the learned structure transfers to completely unseen genes.

In text, inductive transfer fails on both benchmarks tested. A model trained on one set of passage co-occurrences does not improve retrieval for unseen co-occurrences, even when individual passages overlap between splits \citep{dury2026aar}.

One interpretation consistent with this pattern is that text co-occurrence is contingent: the fact that a passage about Tarantino co-occurs with a passage about Knoxville reflects an editorial decision, not a structural regularity. No feature of the embeddings predicts this specific pairing. Protein interactions, by contrast, reflect physical constraints---binding interfaces, complex stoichiometry, signalling cascade topologies---that may produce recurring regularities in the perturbation profiles. The MLP can generalise from these regularities to unseen genes.

We note that alternative explanations exist: the difference could reflect embedding dimensionality (50 vs 1024), dataset size, or the structure of the association graph rather than a deep distinction between contingent and physical associations. Disentangling these factors requires further cross-domain comparisons.

\FloatBarrier
\subsection{Conditions for Success}
\label{sec:conditions}

\begin{table}[h]
\centering
\caption{Operating envelope: when does CAL work?}
\label{tab:envelope}
\begin{tabular}{llll}
\toprule
Experiment & Regime & Failure mode & Outcome \\
\midrule
Gene (Replogle) & Latent signal present & --- & +0.390 CB AUC \\
DepMap (random neg.) & Rescued & Negative sampling & +0.377 CB AUC \\
DepMap (in-batch) & Neg.\ sampling failure & In-batch neg.\ from pos.\ cluster & $-$0.517 \\
Drug fingerprints & No latent signal & Embedding uninformative & +0.021 \\
Drug L1000 & Degree confounded & Low entity diversity & Shuffled $>$ real \\
\bottomrule
\end{tabular}
\end{table}

Three conditions emerge from these experiments as practical requirements for CAL to succeed:

\textbf{1.\ The embedding must contain latent associative signal.} The input features must encode information relevant to the association, even if cosine similarity does not surface it. Gene expression profiles encode pathway membership, complex participation, and regulatory relationships---information that relates to physical protein interactions even when direct cosine similarity is low. The MLP extracts this latent mapping. Morgan fingerprints encode local atomic neighbourhoods, which in this setup contained no signal predictive of cellular co-lethality.

Both the Replogle and drug fingerprint experiments had cross-boundary cosine baselines near chance ($\sim$0.53). The difference is that one embedding contains latent signal the MLP can extract and the other does not. A low cosine baseline is necessary but not sufficient.

\textbf{2.\ Negative sampling must match data structure.} When positive pairs cluster in embedding space (DepMap: 98\% cosine $> 0.5$), in-batch negatives are drawn from the same cluster as positives, causing the model to push together what it should separate. Random negatives resolve this by providing negatives from outside the positive cluster.

\textbf{3.\ Entity diversity must be sufficient relative to pair count.} When few entities appear in many pairs (Drug L1000: 2,094 drugs in 200K pairs), the model learns entity frequency rather than pairwise association. This confound is detectable by the shuffled ablation: if the shuffled model matches or exceeds the reference, degree structure is driving the result.

\FloatBarrier
\subsection{Degree-Dependence}
\label{sec:degree}

Two related but distinct findings characterise the relationship between gene interaction degree and CAL performance.

First, the raw association score anti-correlates with degree: Spearman $r = -0.590$ ($p < 10^{-80}$) between STRING degree and CAL association score across all evaluated pairs, while cosine similarity is uncorrelated with degree ($r = 0.007$, $p = 0.32$). This means the MLP assigns lower association scores to pairs involving high-degree genes.

Second, in the cross-boundary regime where CAL provides its unique contribution, the improvement over cosine also favours low-degree genes:

\begin{table}[h]
\centering
\caption{Cross-boundary improvement by degree quintile (Replogle, high confidence). See also Figure~\ref{fig:degree}.}
\label{tab:quintile}
\begin{tabular}{llrrrr}
\toprule
Quintile & Degree range & N pairs & Cosine & Association & $\Delta$ \\
\midrule
Q1 (lowest) & 5--184 & 950 & 0.003 & 0.224 & +0.221 \\
Q5 (highest) & 521--1{,}110 & 933 & $-$0.030 & 0.097 & +0.127 \\
\bottomrule
\end{tabular}
\end{table}

Low-degree genes benefit approximately twice as much as high-degree genes in the cross-boundary regime.

The mechanism is architectural. Contrastive learning with batch size 512 maps each gene toward the centroid of its association neighbourhood. A gene with 5 interaction partners receives a clear training signal---each partner is a distinct target. A gene with 500 partners maps toward a diffuse centroid that represents none of them strongly. This is the same single-point prediction constraint identified in \citet{dury2026pam}.

Evaluation with degree-matched negatives (replacing random negatives with negatives matched to the degree distribution of each positive pair) yields AUC 0.915, slightly higher than random negatives (0.910). The reported gains are not inflated by degree structure in the negative distribution.

The practical implication is that CAL is most useful for genes with focused interaction profiles. Hub genes (degree $> 500$) are already extensively annotated. Low-degree genes---the understudied genes most in need of functional characterisation---are precisely where CAL adds the most signal.

This pairwise advantage for low-degree genes does not contradict the hub-driven cluster composition reported in Section~\ref{sec:clusters}. Hub genes form clusters because their many individually weaker association scores create locally dense neighbourhoods that HDBSCAN detects, even though each individual pairwise score is lower than for low-degree genes.

\begin{figure}[t]
\centering
\includegraphics[width=0.75\textwidth]{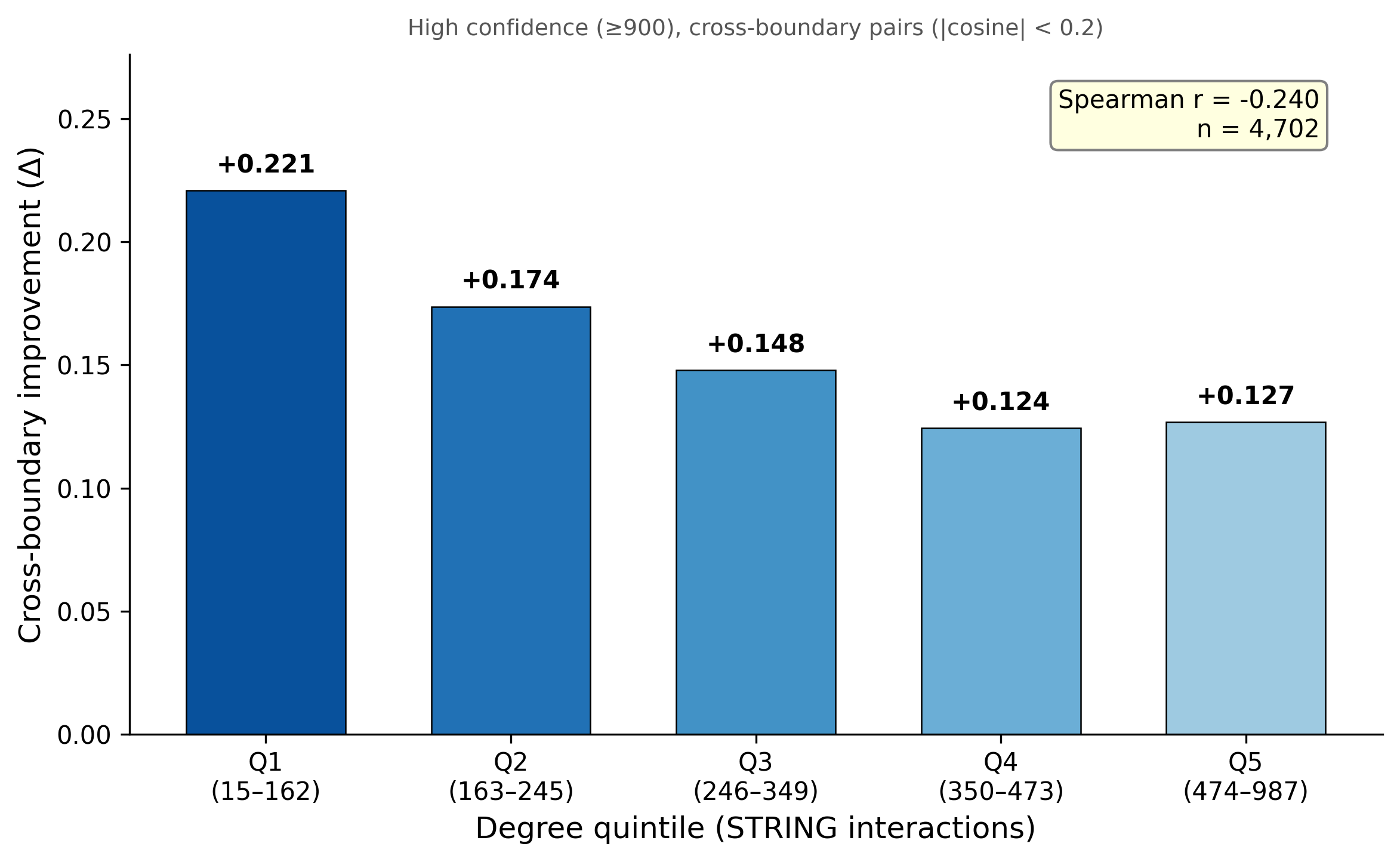}
\caption{Cross-boundary improvement by degree quintile (high confidence $\geq$900, cross-boundary pairs with $|\text{cosine}| < 0.2$). Low-degree genes (Q1) benefit approximately twice as much as high-degree genes (Q5). The trend is approximately monotonic; Q4 and Q5 show near-equal improvement (+0.124 vs +0.127), suggesting the degree effect saturates for high-degree genes. Table~\ref{tab:quintile} shows Q1 and Q5 from the same high-confidence cross-boundary data; the appendix (Table~\ref{tab:degree_full}) uses medium-confidence all-pairs.}
\label{fig:degree}
\end{figure}

\subsection{Illustrative Examples}
\label{sec:examples}

Three gene pairs illustrate the practical value of CAL for recovering functional relationships invisible to expression similarity. These examples are illustrative rather than exhaustive; the degree-dependence analysis (Section~\ref{sec:degree}) provides the systematic view.

\textbf{C7ORF26--INTS1.} C7ORF26 (degree 32) was recently reclassified as Integrator complex subunit INTS15. Expression cosine with INTS1 is 0.074 (indistinguishable from random at mean pairwise cosine 0.088). Association score: 0.393. The STRING experimental evidence score is 994, but the text-mining score is 0 for this pair---the interaction was characterised experimentally before it entered the literature. CAL recovers a recently validated interaction that is invisible to expression similarity and absent from text-mining evidence.

\textbf{SNAPC complex.} All four characterised SNAPC complex subunits (SNAPC1--4) have low interaction degree ($\leq$36). SNAPC4 is the invisible fourth subunit: its expression cosine with the other three ranges from $-$0.087 to 0.167, well within the noise floor. CAL groups all four: SNAPC2--SNAPC4 (cosine 0.167, association 0.471), SNAPC3--SNAPC4 (cosine 0.087, association 0.399), SNAPC1--SNAPC4 (cosine $-$0.087, association 0.361). These subunits form a physical complex with strong experimental evidence ($>$780) but SNAPC4 is invisible to cosine similarity. SNAPC2 at degree 12 is among the lowest-degree genes in the dataset.

\textbf{HSPA5--MANF.} MANF (degree 22) is a relatively understudied endoplasmic reticulum stress protein. Its expression profile is anti-correlated with HSPA5 (cosine $-$0.048). Association score: 0.353---the largest single-pair improvement in the dataset (+0.401). The interaction is experimentally validated (STRING experimental score 805) but the anti-correlated expression profiles would cause any similarity-based method to rank it among the least likely interactions.

\begin{figure}[t]
\centering
\includegraphics[width=\textwidth]{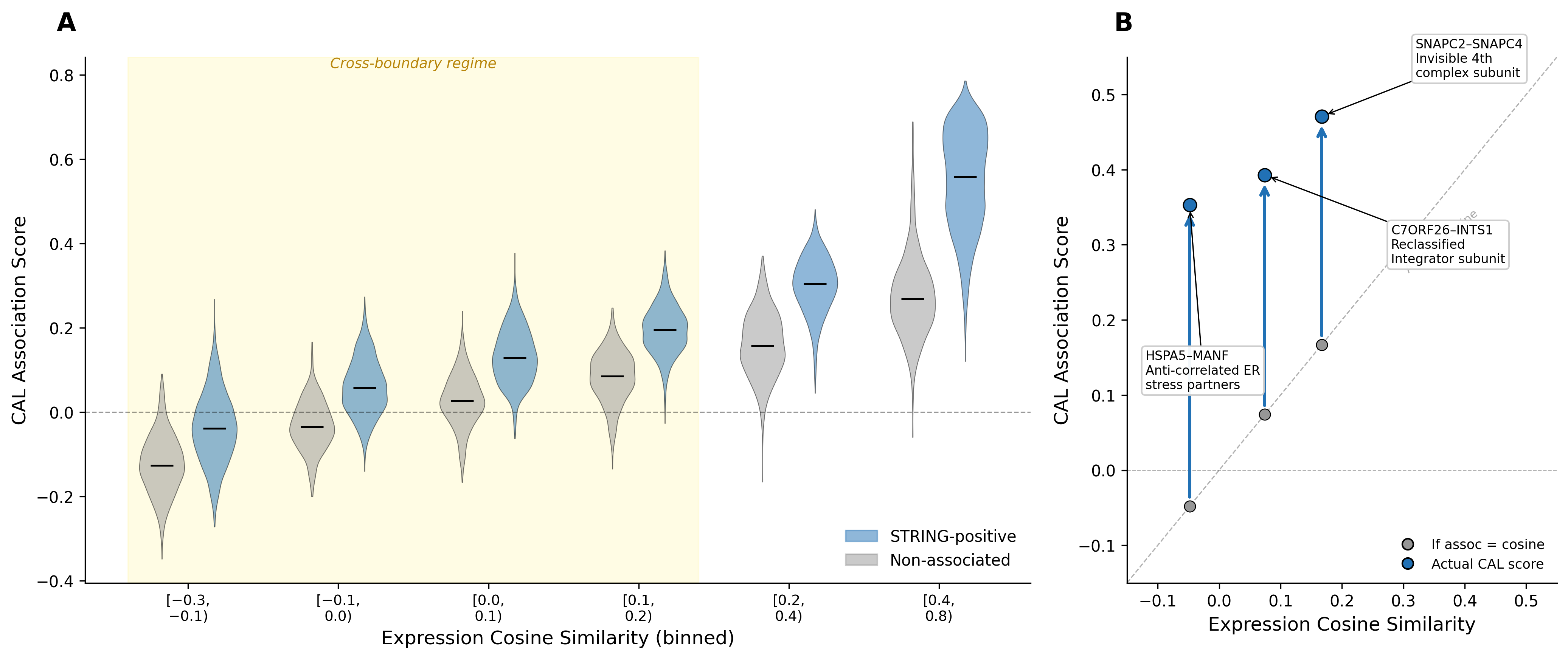}
\caption{(A)~Association score distributions by cosine similarity bin. In the cross-boundary regime (yellow shading), STRING-positive pairs (blue) show clear separation from non-associated pairs (grey). (B)~Three illustrative gene pairs showing CAL lift: each arrow indicates the gain from cosine (grey dot on diagonal) to the actual association score (blue dot).}
\label{fig:examples}
\end{figure}

\FloatBarrier
\section{Cluster Validation}
\label{sec:clusters}

HDBSCAN clustering (min\_cluster\_size=5, min\_samples=3) applied to the association-space distance matrix ($1 - \text{cosine similarity}$ of CAL-transformed embeddings, high-confidence model) identifies 66 clusters containing 838 genes. Expression-space clustering with identical parameters identifies 36 clusters containing 1,088 genes. Twenty-two clusters are PAM-specific: clusters found in association space whose members are either predominantly unclustered in expression space ($>$50\% noise) or span three or more distinct expression-space clusters.

Gene Ontology enrichment analysis (g:Profiler, g\_SCS multiple testing correction across GO:BP, GO:MF, GO:CC, Reactome, and KEGG) classifies each cluster as coherent (corrected $p < 0.05$, $>$50\% gene coverage), partial ($p < 0.05$, $\leq$50\% coverage), or no signal.

\begin{table}[h]
\centering
\caption{Cluster GO enrichment comparison.}
\label{tab:clusters}
\begin{tabular}{lrrrrr}
\toprule
Cluster source & Total & Coherent & Partial & No signal & \% coherent \\
\midrule
PAM-specific (assoc.\ space) & 22 & 22 & 0 & 0 & 100\% \\
Expression-space HDBSCAN & 36 & 35 & 1 & 0 & 97\% \\
Random (matched sizes, 10 rep.) & 34 & $13.8 \pm 1.4$ & --- & --- & 41\% \\
\bottomrule
\end{tabular}
\end{table}

Expression-space clusters show 97\% GO enrichment, confirming that gene expression captures pathway structure---as expected, since co-regulated genes tend to participate in shared biological processes. The random baseline (41\%) establishes that enrichment is non-trivial for both real clusterings. CAL discovers 22 additional coherent functional groupings that expression-space clustering does not recover.

Among the PAM-specific clusters, 19 of 22 (86\%) are hub-driven ($>$50\% of members have $>$50 STRING interactions). Hub genes have diffuse expression signatures because their perturbation affects many pathways simultaneously---exactly the genes whose functional grouping cosine similarity misses. Three non-hub clusters correspond to Golgi vesicle transport, dynactin complex, and mitochondrial nucleoid.

The most dramatic expression-to-association improvements: mRNA splicing (14 genes, expression cosine $0.024 \to$ association cosine 0.325, 13.7$\times$), telomere capping (5 genes, $0.104 \to 0.598$, 5.75$\times$), transcription elongation (16 genes, $0.216 \to 0.671$, 3.11$\times$), and translation initiation (22 genes, $0.274 \to 0.807$, 2.95$\times$).

\FloatBarrier
\section{Discussion}
\label{sec:discussion}

\subsection{Association $\neq$ Similarity Is Cross-Domain}

The primary claim is empirical: the same contrastive MLP architecture, with no modification beyond input dimensionality, learns useful associative structure from both passage co-occurrence in text and protein interactions in biology. The cross-boundary regime---where cosine similarity is uninformative---is where CAL provides its value, and this regime exists in both domains.

\subsection{Domain Differences}

\begin{table}[h]
\centering
\caption{Domain comparison: text vs biology.}
\label{tab:domains}
\begin{tabular}{lll}
\toprule
Aspect & Text (HotpotQA) & Biology (Replogle) \\
\midrule
Embeddings & BGE-large (1024d) & PCA-50 expression \\
Associations & Gold co-occurrence & STRING PPI \\
Cross-boundary gain & +28.5 R@5 (hard Qs) & +0.390 AUC \\
Inductive transfer (node-split) & Fails ($\pm$0.10) & Succeeds (+0.127) \\
Quality vs quantity & More pairs better & Tighter pairs better \\
Cosine contribution & Useful ($\lambda$=0.6 optimal) & Uninformative (assoc-only best) \\
\bottomrule
\end{tabular}
\end{table}

Three mechanistic differences distinguish the domains:

\textbf{Quality over quantity.} In text, more co-occurrence pairs increase association graph coverage and improve performance. In biology, tighter confidence thresholds outperform larger noisy sets, and experimental-only filtering matches the highest combined-score threshold. Low-confidence STRING edges include text-mining and co-expression (similarity signals); high-confidence and experimental-only edges carry stronger physical interaction evidence.

\textbf{Inductive transfer.} Text associations do not transfer. Biological associations partially transfer even to completely unseen genes (+0.127), consistent with the interpretation that physically grounded associations produce embedding-space regularities that the MLP can generalise from. Alternative explanations (embedding dimensionality, graph structure) cannot be ruled out from these experiments alone.

\textbf{Cosine contribution.} The biological embedding space is less informative about protein interactions than text embeddings are about passage co-occurrence. The learned association signal monotonically dominates cosine at every blend weight (Appendix~\ref{app:lambda}), so cosine contributes nothing in this domain. In text, the optimal blend balances both signals.

\subsection{Practical Deployment Guidance}

For practitioners considering CAL on new data:

\begin{enumerate}
    \item \textbf{Check embedding--association independence.} Compute cosine AUC on association labels. If already $> 0.85$, cosine captures most of the signal; CAL may help only at the margin and requires random negatives.
    \item \textbf{Check positive pair cosine distribution.} If $>$50\% of positive pairs have cosine $> 0.5$, use random negatives rather than in-batch negatives.
    \item \textbf{Check entity-to-pair ratio.} If entities appear in $>$50 pairs on average, degree confounding is likely. Run the shuffled ablation early.
    \item \textbf{Prefer high-quality associations over high-quantity.} If association annotations have varying confidence, train on the highest-confidence subset first.
    \item \textbf{Run the shuffled ablation.} If shuffled $\geq$ reference, the model has learned degree structure rather than genuine associations. This is the single most important diagnostic.
\end{enumerate}

\subsection{Connection to the PAM Programme}

The text and biological experiments together provide converging evidence for PAM's central claim: that co-occurrence produces associative structure that cosine similarity cannot recover. The concept discovery paper \citep{dury2026concepts} showed this at corpus scale in text, discovering narrative function rather than topic. The present work shows it holds in a domain with independently derived ground truth.

The inductive transfer asymmetry offers a refinement to PAM's original framing, which emphasised that association should be tied to experienced co-occurrences \citep{dury2026pam}. The biological results suggest this may be too strong when associations reflect stable physical constraints. Inductive transfer success may serve as an empirical signature of physically grounded association structure---a prediction testable in further cross-domain work.

\section{Limitations}
\label{sec:limitations}

Several limitations scope the claims made in this work.

First, the gene experiments use transductive evaluation: the association model is trained on pairs drawn from the same dataset on which it is evaluated. The inductive ablations (edge-split and node-split) partially address this, but full inductive evaluation on an independent gene interaction database (e.g., BioGRID, IntAct) was not performed.

Second, STRING associations are not a pure experimental ground truth. The combined score aggregates multiple evidence channels including text-mining and co-expression. We validate against the experimental channel separately (Section~\ref{sec:quality}), but the primary results use combined scores.

Third, all gene results use a single embedding choice (PCA-50 of CRISPRi expression profiles). Whether CAL succeeds with other dimensionalities (PCA-100, PCA-128) or embedding strategies (autoencoders, learned representations) is untested. PCA-50 captures approximately 50\% of variance, which means potentially informative signal is discarded.

Fourth, the paper compares mainly against cosine similarity and internal ablations rather than dedicated PPI prediction or metric-learning baselines. This is appropriate for testing the association $\neq$ similarity principle but does not establish CAL's practical utility relative to existing biological tools.

Fifth, evaluation choices---the cross-boundary threshold ($|\text{cos}| < 0.2$), negative sampling (random, 5$\times$), and pair deduplication---may affect reported gains. Cross-boundary AUC is robust to threshold choice (Appendix~\ref{app:cb}), and degree-matched negatives produce slightly higher AUC than random negatives (0.915 vs 0.910), but other evaluation choices were not systematically varied.

Sixth, the degree-dependence finding and the conditions-for-success framework are derived from a small number of experiments. They represent patterns observed across four datasets, not validated general principles.

Seventh, the illustrative gene pairs (Section~\ref{sec:examples}) are selected for dramatic improvement. The degree-dependence analysis (Section~\ref{sec:degree}) provides the systematic view.

Eighth, cluster validation relies on Gene Ontology enrichment, which may be biased toward well-studied pathways. Expression-space clusters show 97\% coherence under the same criteria, establishing a high baseline that the 100\% PAM-specific enrichment rate should be evaluated against.

Ninth, all training uses single runs at each seed. The bootstrap confidence intervals capture evaluation variance, and the three-seed analysis shows near-zero training variance (SD $< 0.001$), but the seed analysis was performed only on the high-confidence model.

\section{Future Work}
\label{sec:future}

\textbf{Independent validation datasets.} Evaluate CAL trained on STRING against independent interaction databases (BioGRID, IntAct) or functional genomics assays (CRISPRi-FlowFISH enhancer--gene links).

\textbf{Node-split at scale.} The current node-split uses Replogle (2,285 genes). Genome-wide CRISPRi datasets ($>$10,000 perturbations) would test whether cold-start generalisation holds at larger scale and whether the degree-dependence pattern persists.

\textbf{Additional cross-domain tests.} Legal case law (CourtListener corpus), where citation structure provides co-occurrence annotations, is a natural next test case. Citations may be more structurally grounded than text passage co-occurrence but less physically grounded than protein interactions, testing whether inductive transfer success correlates with association groundedness.

\textbf{Addressing degree confounding.} Degree-aware negative sampling or degree-normalised contrastive losses may mitigate the anti-correlation between interaction degree and CAL improvement, extending utility to hub genes.

\textbf{Embedding sensitivity.} Systematic comparison of PCA-50, PCA-100, and autoencoder embeddings on the Replogle dataset would establish whether the current results are robust to embedding choice.

\section{Conclusion}
\label{sec:conclusion}

Contrastive Association Learning generalises from text to biology. The same 4-layer MLP, trained with the same contrastive loss on protein interaction annotations, recovers functional gene relationships invisible to expression similarity---achieving cross-boundary AUC of 0.908 where cosine similarity scores 0.518. A second gene dataset confirms the result (cross-boundary AUC 0.947) after a negative sampling correction. Results on the primary dataset are stable across training seeds (SD $< 0.001$) and cross-boundary threshold choices.

Two drug experiments identify the boundaries: CAL requires latent signal in the embedding about the association, sufficient entity diversity, and negative sampling matched to data structure.

Three cross-domain findings emerge. First, inductive transfer succeeds in biology---including to completely unseen genes---where it fails in text, consistent with the interpretation that physically grounded associations are more transferable than contingent co-occurrences. Second, improvement concentrates on low-degree genes where the biological need is greatest. Third, association quality dominates quantity, reversing the text pattern. Taken together, these results indicate that association $\neq$ similarity is a cross-domain phenomenon and provide preliminary conditions for predicting when contrastive association learning will succeed on new data. Code and trained models are available at \url{https://github.com/EridosAI/GeneticCAL}.

\appendix

\section{Lambda Sweep}
\label{app:lambda}

The association signal monotonically dominates cosine at every blend weight. Higher $\lambda$ (more association, less cosine) strictly improves AUC at all three confidence thresholds.

\begin{table}[h]
\centering
\caption{Overall AUC as a function of $\lambda$ (Replogle).}
\label{tab:lambda}
\begin{tabular}{rrrr}
\toprule
$\lambda$ & Low ($\geq$400) & Medium ($\geq$700) & High ($\geq$900) \\
\midrule
0.0 (cosine only) & 0.570 & 0.644 & 0.692 \\
0.1 & 0.575 & 0.652 & 0.701 \\
0.2 & 0.582 & 0.661 & 0.711 \\
0.3 & 0.589 & 0.672 & 0.724 \\
0.4 & 0.599 & 0.685 & 0.739 \\
0.5 & 0.612 & 0.703 & 0.757 \\
0.6 & 0.630 & 0.725 & 0.780 \\
0.7 & 0.655 & 0.754 & 0.809 \\
0.8 & 0.691 & 0.793 & 0.845 \\
0.9 & 0.744 & 0.840 & 0.883 \\
1.0 (assoc only) & 0.800 & 0.877 & 0.910 \\
\bottomrule
\end{tabular}
\end{table}

This monotonic relationship eliminates the need for $\lambda$ selection on the evaluation set---the optimal choice is always $\lambda = 1.0$ (association-only). This contrasts with text retrieval, where the optimal $\lambda = 0.6$ reflects genuine complementarity between cosine similarity and association \citep{dury2026aar}.

\begin{table}[h]
\centering
\caption{Cross-boundary AUC as a function of $\lambda$ (medium confidence).}
\label{tab:lambda_cb}
\begin{tabular}{rr}
\toprule
$\lambda$ & CB AUC \\
\midrule
0.0 & 0.534 \\
0.1 & 0.566 \\
0.2 & 0.603 \\
0.3 & 0.644 \\
0.4 & 0.689 \\
0.5 & 0.737 \\
0.6 & 0.782 \\
0.7 & 0.819 \\
0.8 & 0.843 \\
0.9 & 0.856 \\
1.0 (assoc only) & 0.861 \\
\bottomrule
\end{tabular}
\end{table}

\section{Cross-Boundary Threshold Sensitivity}
\label{app:cb}

\begin{table}[h]
\centering
\caption{CAL AUC at varying cross-boundary thresholds (high confidence $\geq$900). See also Figure~\ref{fig:cb_sensitivity}.}
\label{tab:cb_sensitivity}
\begin{tabular}{rrrrrr}
\toprule
$|$cos$|$ threshold & Pos.\ pairs & Neg.\ pairs & Cosine AUC & CAL AUC & $\Delta$ \\
\midrule
$< 0.30$ & 6{,}805 & 27{,}671 & 0.540 & 0.907 & +0.367 \\
$< 0.20$ & 4{,}706 & 19{,}580 & 0.518 & 0.908 & +0.390 \\
$< 0.15$ & 3{,}643 & 15{,}109 & 0.516 & 0.909 & +0.393 \\
$< 0.10$ & 2{,}368 & 10{,}327 & 0.510 & 0.914 & +0.404 \\
$< 0.05$ & 1{,}211 & 5{,}293 & 0.491 & 0.915 & +0.424 \\
\bottomrule
\end{tabular}
\end{table}

CAL AUC increases monotonically as the cross-boundary threshold tightens, while cosine drops toward chance. The gap widens from +0.37 to +0.42, confirming that CAL's advantage grows in the regime where similarity is least informative.

\begin{figure}[h]
\centering
\includegraphics[width=0.75\textwidth]{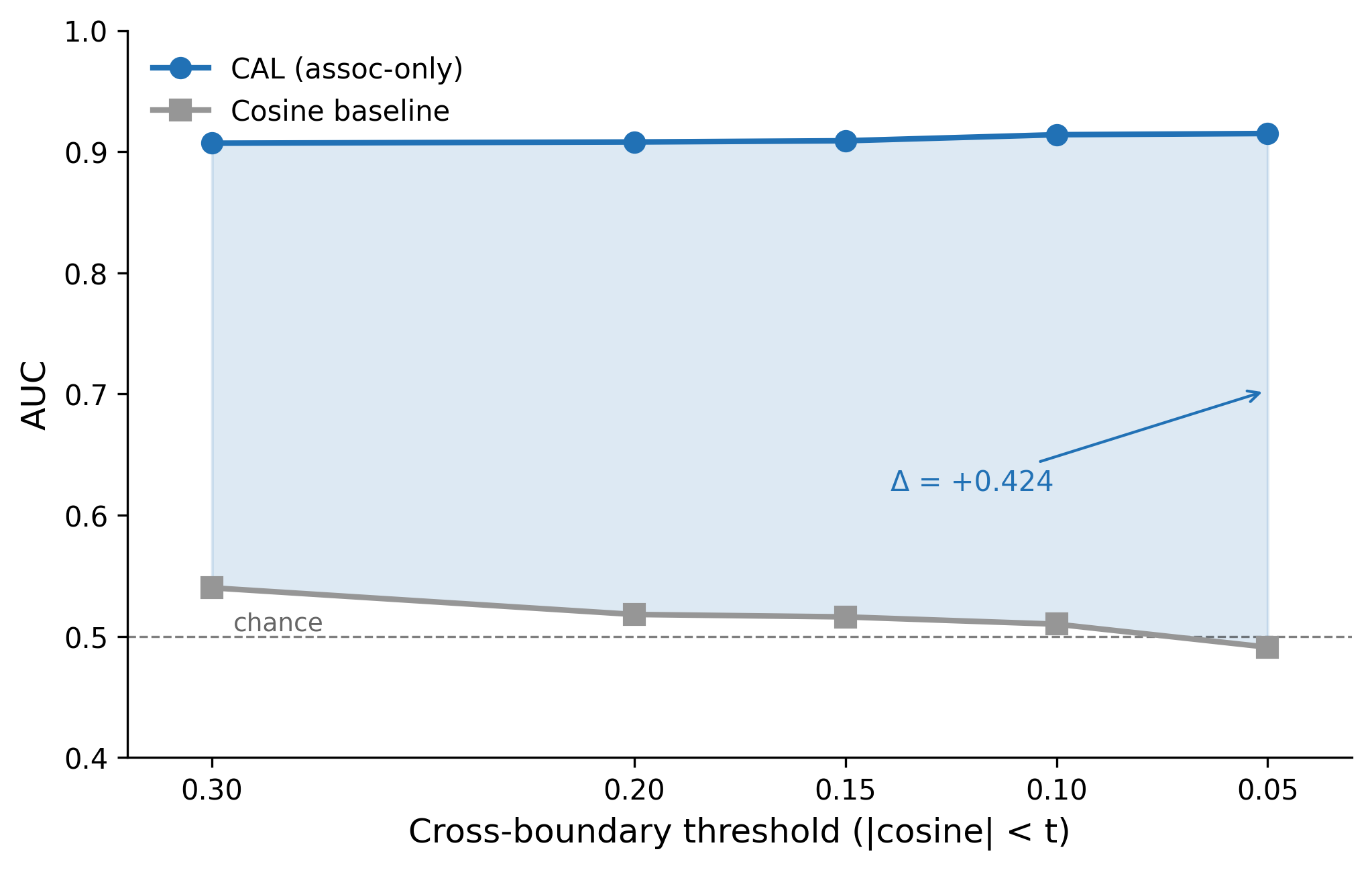}
\caption{Cross-boundary sensitivity analysis. As the threshold tightens (left to right), cosine AUC declines toward chance while CAL AUC remains stable or slightly increases. Shaded region shows the growing gap.}
\label{fig:cb_sensitivity}
\end{figure}

\section{DepMap Detailed Results}
\label{app:depmap}

\begin{table}[h]
\centering
\caption{DepMap in-batch training across pair counts.}
\label{tab:depmap_detail}
\begin{tabular}{rrrr}
\toprule
N pairs & CAL AUC & Cosine AUC & $\Delta$ \\
\midrule
25K & 0.336 & 0.924 & $-$0.588 \\
50K & 0.341 & 0.916 & $-$0.576 \\
100K & 0.279 & 0.908 & $-$0.628 \\
200K & 0.379 & 0.896 & $-$0.517 \\
\bottomrule
\end{tabular}
\end{table}

Full ablation results (in-batch): shuffled 0.442, similar positives 0.710, inductive 0.235. Random negatives cross-boundary at $|\text{cos}| < 0.1$: CAL 0.946 vs cosine 0.539.

STRING external validation (random negatives model): overall CAL 0.616--0.695 vs cosine 0.651--0.712; cross-boundary CAL 0.537--0.552 vs cosine 0.516--0.522.

\section{Drug Experiment Details}
\label{app:drugs}

\subsection{Morgan Fingerprints (Experiment 3)}

\begin{table}[h]
\centering
\caption{Drug structure training results.}
\label{tab:drug_morgan}
\begin{tabular}{rrrr}
\toprule
N pairs & CAL AUC & Cosine AUC & $\Delta$ \\
\midrule
25K & 0.305 & 0.565 & $-$0.260 \\
200K & 0.554 & 0.534 & +0.021 \\
Shuffled & 0.636 & --- & $>$ real \\
\bottomrule
\end{tabular}
\end{table}

Validation against mechanism of action (CAL 0.553 vs cosine 0.716) and drug target (CAL 0.582 vs cosine 0.772). A pivot to RDKit molecular descriptors also failed (CAL 0.482 vs cosine 0.514 at 200K pairs).

\subsection{L1000 Transcriptional Signatures (Experiment 4)}

\begin{table}[h]
\centering
\caption{Drug L1000 training results (200K pairs).}
\label{tab:drug_l1000}
\begin{tabular}{lrrrr}
\toprule
Setting & CAL AUC & CB AUC & Shuffled AUC & Shuffled CB \\
\midrule
In-batch neg. & 0.674 & 0.679 & --- & --- \\
Random neg. & 0.850 & 0.842 & 0.909 & 0.908 \\
\bottomrule
\end{tabular}
\end{table}

Degree-controlled analysis: reference $\rho = 0.649$, shuffled $\rho = 0.917$. Lowest-degree drugs (Q0, 420 pairs): reference 0.904 vs shuffled 0.813 (+0.091).

\section{Degree-Dependence Full Results}
\label{app:degree}

\begin{table}[h]
\centering
\caption{Full degree quintile analysis (Replogle, medium confidence)---all pairs.}
\label{tab:degree_full}
\begin{tabular}{llrrr}
\toprule
Quintile & Degree range & Cosine & Association & $\Delta$ \\
\midrule
Q1 (lowest) & 5--184 & 0.364 & 0.252 & $-$0.112 \\
Q2 & 185--270 & 0.431 & 0.215 & $-$0.216 \\
Q3 & 271--386 & 0.437 & 0.180 & $-$0.257 \\
Q4 & 387--520 & 0.380 & 0.143 & $-$0.238 \\
Q5 (highest) & 521--1{,}110 & 0.336 & 0.108 & $-$0.228 \\
\bottomrule
\end{tabular}
\end{table}

Note: The all-pairs analysis shows cosine outperforming association at every quintile because it includes high-cosine pairs where cosine similarity already captures the interaction. The cross-boundary analysis (Table~\ref{tab:quintile} in main text) isolates the regime where CAL provides its unique contribution.

\section{Cluster Validation Details}
\label{app:clusters}

Hub vs non-hub analysis of 22 PAM-specific clusters:

\begin{table}[h]
\centering
\caption{Hub vs non-hub cluster properties.}
\label{tab:hub}
\begin{tabular}{lrrr}
\toprule
 & Expr.\ cosine & Assoc.\ cosine & Ratio \\
\midrule
Hub clusters ($n$=19) & $0.407 \pm 0.181$ & $0.692 \pm 0.154$ & 1.70$\times$ \\
Non-hub clusters ($n$=3) & $0.728 \pm 0.062$ & $0.587 \pm 0.053$ & 0.81$\times$ \\
\bottomrule
\end{tabular}
\end{table}

Non-hub clusters already have high expression similarity ($\sim$0.73)---they are PAM-specific only in the clustering sense (found in association space but not expression space at matched HDBSCAN parameters), not because CAL found hidden relationships.

The apparent tension between pairwise degree-dependence (Section~\ref{sec:degree}: low-degree genes benefit most) and cluster composition (86\% hub-driven) reflects different levels of analysis. Pairwise improvement measures how much CAL boosts individual interaction scores relative to cosine. Cluster formation depends on the density of association-space neighbours, which is higher for hub genes with many interactions. Hub genes form clusters in association space because their many (individually weaker) association scores still create locally dense regions that HDBSCAN detects, even though each individual pairwise score is lower than for low-degree genes.

\section{Training Details}
\label{app:training}

\subsection{Replogle Main Model (high confidence $\geq$900)}

\begin{itemize}
    \item \textbf{Data:} 2,285 gene perturbations from K562 essential CRISPRi Perturb-seq \citep{replogle2022mapping}. PCA-50, L2-normalised. 23,268 STRING pairs at combined score $\geq$900.
    \item \textbf{Architecture:} 4-layer MLP with learned residual. Linear(50$\to$1024) $\to$ LN $\to$ GELU $\to$ Linear(1024$\to$1024) $\to$ LN $\to$ GELU $\to$ Linear(1024$\to$1024) $\to$ LN $\to$ GELU $\to$ Linear(1024$\to$50) $\to$ LN. Output: $f(\mathbf{x}) = \text{normalize}(\alpha \cdot \mathbf{x} + (1-\alpha) \cdot \text{MLP}(\mathbf{x}))$, $\alpha = \sigma(\text{learned scalar})$. Total: 2,208,919 parameters.
    \item \textbf{Training:} Symmetric InfoNCE, batch size 512, temperature 0.05, AdamW (lr=$3 \times 10^{-4}$, weight\_decay=$1 \times 10^{-4}$), CosineAnnealingLR ($T_{\max}$=100), 100 epochs, seed 42.
    \item \textbf{Scoring at evaluation:} Half-transformed: $\text{score}(A,B) = 0.5 \times (f(\mathbf{e}_A) \cdot \mathbf{e}_B + f(\mathbf{e}_B) \cdot \mathbf{e}_A)$.
    \item \textbf{Negatives at evaluation:} 50,000 random non-positive pairs ($5\times$ positives, capped), seed 42, deduplicated in both directions.
    \item \textbf{Training accuracy:} 5.2\% (chance = 1/512 = 0.2\%).
    \item \textbf{Learned alpha:} 0.490 (near-equal residual/MLP blend).
    \item \textbf{Hardware:} Single CUDA GPU. Training time: $\sim$3 minutes per threshold.
\end{itemize}

\subsection{DepMap Random-Negatives Model}

Differences from Replogle: input dim 100 (PCA-100), 200K co-essentiality pairs, 200K evaluation negatives, both-transformed scoring in original evaluation ($f(\mathbf{A}) \cdot f(\mathbf{B})$). All other hyperparameters identical.

The DepMap evaluation used both-transformed scoring ($f(\mathbf{e}_A) \cdot f(\mathbf{e}_B)$) while Replogle used half-transformed scoring (Section~\ref{sec:evaluation}). Both methods produce qualitatively identical patterns; the choice affects absolute values but not the direction of results. DepMap numbers in the main text use the original both-transformed scoring.

\subsection{Ablation Parameters}

All ablations use identical architecture and hyperparameters. Only the training data differs: shuffled (random permutation of one pair column), similar positives (highest-cosine pairs), random negatives (uniform random instead of in-batch), inductive (subset of pairs).

\bibliographystyle{plainnat}
\bibliography{cal_references}

\end{document}